\documentclass[preprint,review,12pt]{elsarticle}

\usepackage{amssymb}
\usepackage{graphicx}
\usepackage{subfigure}
\usepackage{multicol}
\usepackage{multirow}
\usepackage{url}
\usepackage{algorithm}
\usepackage{algorithmic}
\usepackage{rotating}
\usepackage{epstopdf}

\usepackage{enumitem}
\setenumerate[1]{itemsep=0pt,partopsep=0pt,parsep=\parskip,topsep=5pt}
\setitemize[1]{itemsep=0pt,partopsep=0pt,parsep=\parskip,topsep=0pt}
\setdescription{itemsep=0pt,partopsep=0pt,parsep=\parskip,topsep=0pt}

\usepackage{amsmath}
\usepackage{multirow}
\usepackage{bbm}
\allowdisplaybreaks[4]

\usepackage{algorithm}
\usepackage{algorithmic}

\begin{document}
	
	\begin{frontmatter}
		
		\title{Identity-Aware CycleGAN for Face Photo-Sketch Synthesis and Recognition}
		
		\author{Yuke Fang}
		\author{Jiani Hu}
		\author{Weihong Deng\corref{cor1}}
		\ead{whdeng@bupt.edu.cn}
		
		\cortext[cor1]{Corresponding author. Tel:+86 10 62283059 Fax: +86 10-62285019}
		
		\address{School of Information and Communication Engineering, Beijing University \\of Posts and
			Telecommunications, Beijing, 100876, China}
		
		\begin{abstract}
			
			Face photo-sketch synthesis and recognition has many applications in digital entertainment and law enforcement. Recently, generative adversarial networks (GANs) based methods have significantly improved the quality of image synthesis, but they have not explicitly considered the purpose of recognition. In this paper, we first propose an Identity-Aware CycleGAN (IACycleGAN) model that applies a new perceptual loss to supervise the image generation network. It improves CycleGAN on photo-sketch synthesis by paying more attention to the synthesis of key facial regions, such as eyes and nose, which are important for identity recognition. 
			Furthermore, we develop a mutual optimization procedure between the synthesis model and the recognition model, which iteratively synthesizes better images by IACycleGAN and enhances the recognition model by the triplet loss of the generated and real samples. 
			Extensive experiments are performed on both photo-to-sketch and sketch-to-photo tasks using the widely used CUFS and CUFSF databases. The results show that the proposed method performs better than several state-of-the-art methods in terms of both synthetic image quality and photo-sketch recognition accuracy.
			
		\end{abstract}
		
	
	\end{frontmatter}

	\section{Introduction}
	
		Face photo-sketch synthesis (FPSS) has received significant interest for its various applications in both social entertainment and police enforcement \cite{Wang2009Face}. 
		For digital entertainment, people want to obtain their own face sketches immediately by taking photos to be shared with others on social media.
		For law enforcement, a typical application is the automatic matching of one suspect’s facial image to photos in the police's criminal face databases. However, in many situations, the facial photo of a potential criminal may not be available because only poor-quality images under partial occlusion are captured. People have found that face sketches can be used as a substitute in searching for a suspect by constructing a mapping between a facial photo and a sketch. 
		The common solution to this issue is performing FPSS and then implementing face photo-sketch recognition (FPSR) in a single domain. 
		In this aspect, the tasks of FPSS and FPSR can be studied together.
	
		There are two pathways for synthesis-based methods in implementing photo-sketch recognition: one is matching in the sketch domain by transforming photo to sketch; the other is transforming sketch to photo and then matching in the photo domain \cite{Wang2009Face}.
		Based on model construction techniques, conventional FPSS methods can be divided into three classes \cite{Wang2013Transductive}: subspace learning-based method such as LLE \cite{Liu2005A}, sparse representation-based method such as SFS \cite{Gao2012Face} and Bayesian inference-based method such as MRF \cite{Wang2009Face} and MWF \cite{Zhou2012Markov}.
		However, those FPSS methods mostly concentrate on making synthesized images consistent with the original images in texture, which may cause information loss for face recognition \cite{Wang2013Transductive}. 
		
		Recently, with the generative adversarial networks (GANs) \cite{Goodfellow2014Generative} have become prevalent, considerable progress has been made in the image-to-image translation tasks.
		For a typical GAN model, the discriminator aims to decide whether the given inputs are fake or real, while the generator learns to generate sharper and more realistic samples that are indistinguishable from the real samples.
		For example, Pix2Pix \cite{Isola2017Image} accomplishes the image style translation task in a supervised manner using conditional GAN.
		To alleviate the difficulty of obtaining image pairs, CycleGAN \cite{Zhu2017Unpaired} preserves key attributes between the inputs and the translated images by utilizing a cycle consistency loss. However, when applied to the FPSS task, all these frameworks are only capable of learning the relations between two domains, where its discriminator focuses only on the differences between the photos and sketches but does not consider any specific optimization for the recognition purpose.
		
		\begin{figure}[ht]
			\centering
			\includegraphics[scale=0.6]{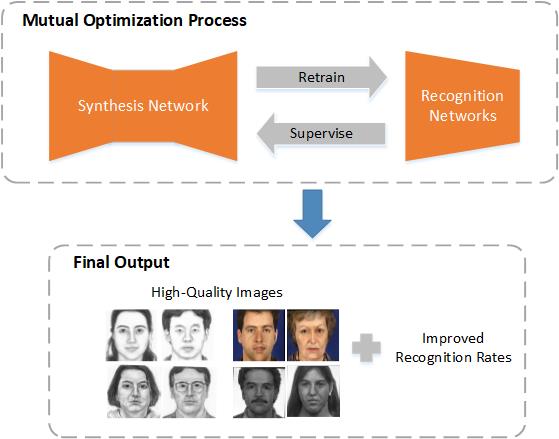}
			\caption{Our mutual optimization process is composed of two parts: synthesis network and recognition networks (respectively for photo and sketch). The generated image dataset from the synthesis network can be used to retrain the recognition networks. And the new recognition model can be considered as a supervision unit to the image generation process in the synthesis network (IACycleGAN), which generates new datasets to construct next optimization cycle. When optimization balances, high-quality synthesized images and improved recognition rates can be obtained simultaneously for both FPSS and FPSR.}
			\label{fig1}
		\end{figure}
		
		To address this problem, every facial image must contain its own "identity-specific information" \cite{Zhang2019Dual}. While Huang \textit{et al.} \cite{Huang2017Face} apply the identity preservation between the input face and the translated face for the image-video translation, we suggest to supervise the relationship between the real target and the fake image using two recognition networks.
		In this paper, we propose an Identity-Aware Cycle Generative Adversarial Network (IACycleGAN) that integrates the recognition model into the cross-domain image generation process. 
		As demonstrated in Figure \ref{fig1}, the recognition networks, first fine-tuned by the generated sketch or photo dataset from the basic synthesis network (CycleGAN), provide a supervision to regularize the performance of the synthesis network (IACycleGAN) using a perceptual loss \cite{Johnson2016Perceptual}.
		Then, to enhance the recognition ability for FPSR, the synthesized images from IACycleGAN can be used to retrain the recognition networks. 
		Similarly, the newly fine-tuned recognition model can be devoted to the synthesis process once again.
		Finally, a mutual cyclic optimization process is formed between the recognition model and the synthesis model, from which work on FPSS and FPSR can be constantly improved.	
			
		In summary, our main contributions include the following:
		\begin{enumerate}
		\item First, we propose IACycleGAN, a robust generative adversarial network that addresses the FPSS problem, from which a perceptual loss is introduced to take the face identity information into consideration between the real and synthesized images. This new loss makes our synthesis network focus on high-level features for identity recognition, rather than the average pixel features of the whole image.
		\item Second, we designed a feedback training method to further enhance the deep recognition model. On the one hand, IACycleGAN supervises the image generation process by the deep recognizer. On the other hand, the generated images from the synthesis network can be used to retrain the recognition model. Experimental results show that our mutual optimization method indeed improves the final recognition accuracy.
		\end{enumerate}
	
		Extensive experiments are performed on two widely used datasets, CUFS and CUFSF, by evaluating both the image quality of the synthesized images and photo-sketch recognition accuracy. The results suggest that the proposed method is effective in improving both sketch-matching and photo-matching based recognition, achieving a much better result compared to state-of-the-art synthesis-based recognition methods such as LR \cite{Wang2017Data-driven}, GAN \cite{Isola2017Image}, DR-GAN \cite{Tran2017Disentangled} and Dual-Transfer \cite{Zhang2019Dual}. We also evaluated the fusion of these two matching methods, which can further improve the recognition results. 
		
		The remainder of the paper is organized as follows. A review of existing FPSS methods and image-to-image translation is presented in Section \ref{Related}. The synthesis network, i.e. IACycleGAN, is introduced in Section \ref{Synthesis Network}. The mutual optimization process between the synthesis model and the recognition model is explained in Section \ref{Recognition Network}. 
		Our detailed experimental results and analyses are presented in Section \ref{Experiments}. And Section \ref{Conclusion} draw conclusions about our work.

	\section{Related Work} \label{Related}
		FPSS is considered as a typical image-to-image translation problem. In this section, the development of these two fields is discussed in detail.
		
		\subsection{Face Photo-Sketch Synthesis} \label{Face Photo-Sketch Synthesis}
			
			Existing FPSS approaches can be classified into 4 categories \cite{Wang2013Transductive}: subspace learning, sparse representation, Bayesian inference and deep-learning-based methods, where the former three categories belong to data-driven method, and the last one belongs to model-driven method \cite{Akram2018A}.
			
			Subspace-learning-based methods mainly include linear subspace-based methods, such as principal component analysis (PCA) \cite{Jolliffe2011Principal}, and nonlinear subspace methods such as local linear embedding (LLE) \cite{Roweis2000Nonlinear}. Tang and Wang \cite{Tang2002Face} assume that the target sketch can be synthesized by a linear combination of training sketches, similar to the relationship within photos. Later, inspired by LLE \cite{Roweis2000Nonlinear}, Liu \textit{et al.} \cite{Liu2005A} proposed a patch-based synthesis method that aims to synthesize the target sketch patch from several training sketch patches according to the photos'. 
			In recent years, sparse-representation-based methods have been widely used in FPSS tasks. Chang \textit{et al.} \cite{Chang2010Face} introduce a sparse coding technique for feature extraction, assuming that the face photo and corresponding sketch have the same sparse representation. To further consider high-frequency information, Gao \textit{et al.} \cite{Gao2012Face} propose a two-step process to improve the quality of synthesized sketches, during which the mapping between the high-frequency information of the photo-sketch patches is learned using the sparse coding technique.
			Bayesian inference-based methods explore such an area in a different way by using embedded hidden Markov model (E-HMM) or Markov random field (MRF). 
			Wang and Tang \cite{Wang2009Face} employ a multi-scale MRF model for FPSS problem to learn the relations among neighboring image patches. Zhou \textit{et al.} \cite{Zhou2012Markov} improve the basic MRF method in \cite{Wang2009Face} by proposing a weighted MRF algorithm, called Markov weight fields (MWF). Wang \textit{et al.} \cite{Wang2013Transductive} propose a novel face sketch synthesis method (TFSPS) to improve the robustness of MWF based on transductive learning. 
			Those Bayesian inference-based methods pay special attention to the relations among target image patches in the corresponding domain but may overlook the domain-to-domain transformations, which may lead to high computation costs in practical use.
			
			The deep-learning-based method actually focuses on model-driven strategy, where the mapping function is learned in advance and then applied to execute the transformation. 
			Zhang \textit{et al.} \cite{Zhang2015End-to-End} propose an end-to-end photo-sketch generation model based on a 7-layer fully convolutional network (FCN) with a joint generative-discriminative optimization process. However, the synthesized images still produce blurring, with less facial texture information obtained.
			With GAN attracting increased attention, the development of FPSS was spawned. Wang \textit{et al.} \cite{Wang2018Back} perform photo-to-sketch synthesis with a GAN-based method and demonstrate high effectiveness; however, sketch-to-photo synthesis is not considered.
			Recently, Wang \textit{et al.} \cite{Wang2018High} applied the multi-adversarial idea to CycleGAN to generate high-resolution images.
			For sketch-to-photo synthesis, Kazemi \textit{et al}. \cite{Kazemi2018Unsupervised} employ an additional geometry-discriminator to distinguish based on high-level facial features.
			However, these approaches have only focused on adapting the architecture of GAN for an improvement on synthesizing images. Instead, our framework additionally adopts a mutual optimization process by exploring identity information of images and utilizing the relation between FPSS and FPSR.
			Zhang \textit{et al.} \cite{Zhang2019Dual} propose a dual-transfer FPSS framework composed of an inter-domain transfer process and an intra-domain transfer process. They get the transferred image from the combination of two transfer results, while we transform images between sketch and photo directly through a synthesis model.
		
		\subsection{Image-to-Image Translation} \label{Image-to-Image Translation}
			Many traditional algorithms have been used to address FPSS problem and achieved effectiveness results. However, these methods still have limitations in image conversion, since they may not be able to learn high-level features of target images and only address certain tasks that are generally single purpose.
			With the success of CNNs, many studies address image translation problems effectively. However, in some respects, traditional CNN methods may lead to blurry and conservative results because the only goal for a CNN model is to minimize the loss function, which is defined and constructed artificially such as in \cite{Zhang2015End-to-End}. Under this training strategy, all reasonable outputs are averaged; this may cause indistinct outputs when only minimizing the predefined loss. 
			
			Considering these consequences, a more automated mechanism for training is needed. Generative networks, such as GANs \cite{Goodfellow2014Generative} and VAEs \cite{Kingma2013Auto}, have become popular and have achieved great results. Their fundamental objective is to train a model in which output images have the same distributions as the ground truth but are separated from other domain. 
			For image translation tasks, GANs have produced more impressive results due to their strong generative and discriminative ability compared to other generative networks.
			However, the selection, design and training process of the GANs are also highly uncertain, directly producing many difficulties. 
			Specifically, the simple training requirements of GANs result in large degree of freedom with no pre-modelling necessary; thus, it may suffer performance losses when there are more pixels. Based on this concept, some explicit external information, such as extra inputs by users and category information, can then be added to the original GAN model to enhance its stability. 
			
			Isola \textit{et al.} \cite{Isola2017Image} propose conditional GAN (cGAN), whose generative process is constrained by extra information such as images, labels and tags. In other words, hints are given during training and play a guiding role in the data generation process for the generator. In addition, traditional GANs lack guidance, which can also be seen as a mode distortion problem since the only goal is to optimize an asymmetric KL divergence. 
			Zhu \textit{et al.} later proposed CycleGAN \cite{Zhu2017Unpaired}, which showed promising results on unsupervised image-to-image translation tasks based on two generators and two discriminators using unpaired image data. 
			In addition, CycleGAN adds a new cycle consistency loss to the original loss function, which provides extra supervision between the input and cyclic output. However, facial content from CycleGAN cannot be well preserved because of the weak content constraint. Inspired by dual learning, Yi \textit{et al.} \cite{Yi2018DualGAN} propose Dual-GAN with a similar unpaired training mechanism based on unsupervised performance.
			More recently, Mo \textit{et al.} \cite{Mo2018InstaGAN} design an instance-aware GAN (InstaGAN) to incorporate the instance information and improve multi-instance transfiguration, which shows no advantages in handling our problem of single face image translation.

	\section{Synthesis Network - IACycleGAN} \label{Synthesis Network}
		
		In this section, we first illustrate the architecture of IACycleGAN and then introduce the objective function for training this model.
		
		\subsection{Framework}
		
			\begin{figure}[h]
				\centering
				\includegraphics[scale=0.6]{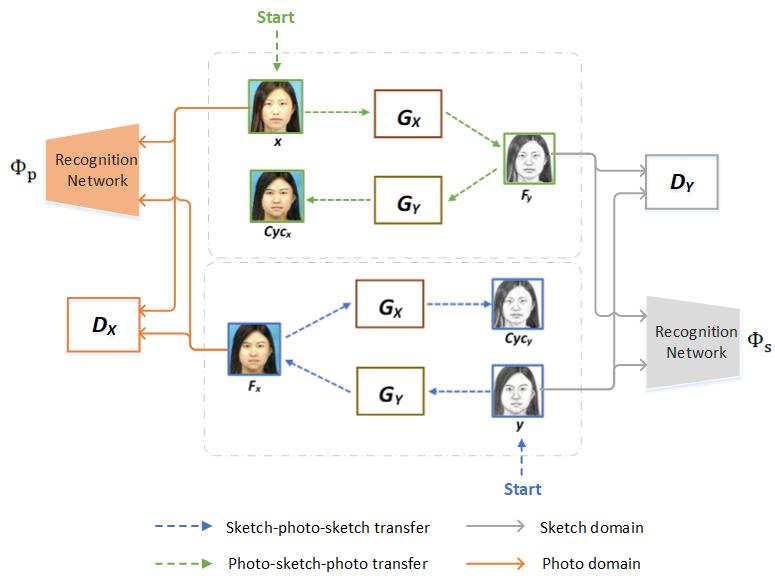}
				\caption{Framework of the IACycleGAN synthesis network. Given a source photo and a sketch reference, our model simultaneously learns a photo-to-sketch transfer function $G_X$ and a sketch-to-photo transfer function $G_Y$. The relations of all representations for images are symmetric between two mapping directions in the training process. We use two recognition models to add an extra perceptual supervision, defined as $\Phi_p$ and $\Phi_s$, for photo and sketch recognition, respectively.}
				\label{fig2}
			\end{figure}
		
			Our goal is to learn the mapping functions between the face photo and the sketch, which correspond to two sample spaces $ X $ and $ Y $.
			A dataset $\mathcal{D}$ consisting of a number of photo-sketch pairs is needed. Given face photos $x$ and corresponding face sketches $y$, we want to obtain two generators $G_X$ and $G_Y$. $G_X$ transforms the face photo into a sketch, and $G_Y$ transforms the face sketch into a photo. The whole network framework is shown in Figure \ref{fig2}. The generated fake sketches from $G_X$ are expressed as $F_y$, and the generated fake photos from $G_Y$ are $F_x$. The relationship is expressed as follows:
			\begin{equation}
			F_y=G_X(x), F_x=G_Y(y).
			\end{equation}
			To realize the cyclic process of the full image transformation, the generated fake images are input into the other generator:
			\begin{equation}
			Cyc_x=G_Y(F_y), Cyc_y=G_X(F_x).
			\end{equation}
			And the two discriminators $D_X$ and $D_Y$ for two domains learn to distinguish between the real input images and the fake generated samples.
			
			Since we have paired data in the target datasets, unsupervised training in \cite{Zhu2017Unpaired,Kazemi2018Unsupervised} is not necessary for our work. 
			To learn the identity information of each person, high-level features of facial images are extracted from a pre-trained recognition model that is fine-tuned based on the VGGFace model \cite{Parkhi2015Deep}. 
			With these two basic recognition networks $\Phi_p$ and $\Phi_s$, respectively for photo and sketch, the most prominent facial characteristic and facial structure for identity discrimination can be captured to directly regularize the distance between the real image and the fake image.
			
			Our IACycleGAN is adapted from the CycleGAN architecture \cite{Zhu2017Unpaired}. The generators come from the style transfer network \cite{Johnson2016Perceptual}, which contains two stride-2 convolutions, several residual blocks [16], and two stride-$\frac{1}{2}$ convolutions. For the discriminators, a PatchGAN architecture is adopted that attempts to classify an image at the patch-level \cite{Isola2017Image,Li2016Precomputed}. 
			Tiling artifacts can be alleviated by using a 70$\times$70 Patch-GAN, and fewer parameters are necessary, from which better results are obtained in terms of both visual quality and efficiency compared to other PatchGAN models of different sizes \cite{Isola2017Image}. 
		
		\subsection{Objective Function}
			
			\begin{figure}[ht]
				\centering
				\includegraphics[scale=0.4]{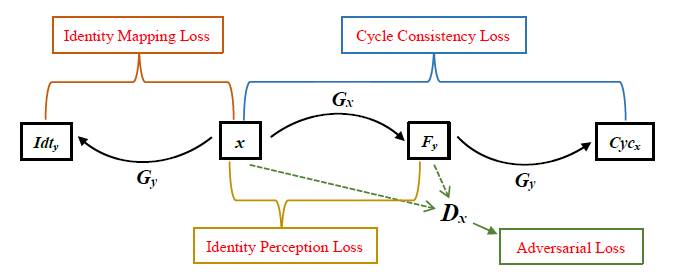}
				\caption{Composition of the final loss function. There are four partial losses: the adversarial loss, cycle consistency loss, identity perception loss and identity mapping loss.}
				\label{fig3}
			\end{figure}
			
			As summarized in Figure \ref{fig3}, the loss function we defined contains four terms: the adversarial loss that matches the distribution of the generated images to the data distribution of the target domain; the cycle consistency loss that prevents the mapping functions $G_X$ and $G_Y$ from contradicting each other; the identity perception loss that aims to preserve identity information based on two recognition networks; and the identity mapping loss that favors to preserve the person texture identity between the input and output.
			
			\textbf{Adversarial Loss.} 
			The adversarial loss is applied in both mapping directions. For the mapping function $G_X: X \rightarrow Y$ and its discriminator $D_Y$, the adversarial loss is expressed as
			\begin{equation}
			\begin{aligned} 
			{\mathcal{L}}_{GAN_X}(G_{X},D_{Y})={{\mathbb{E}}_{y}}[\log D_{Y}(y)] +{{\mathbb{E}}_{x}}[\log (1-D_{Y}(G_{X}(x)))],
			\end{aligned} 
			\end{equation}
			where $ G_X $ aims to generate a fake image $ G_{X}(x) $ that is completely indistinguishable from the images in the target domain, while $D_Y$ tries to distinguish between the real and fake images. And for the mapping function $G_Y: Y \rightarrow X$ and the discriminator $D_X$, we define a similar adversarial loss: $ {\mathcal{L}}_{GAN_Y} $. The generators try to minimize this loss, while discriminators try to maximize it. 
			
			\textbf{Cycle Consistency Loss.} 
			The key to the success of CycleGAN is using the supervision of the cycle consistency loss, which assumes that the generated images can be transformed back to the source domain. Thus, for the images in domain $ X $, we expect $ x \rightarrow G_{X}(x) \rightarrow G_{Y}(G_{X}(x)) \approx x $, and for the images in domain $ Y $, we expect $ y \rightarrow G_{Y}(y) \rightarrow G_{X}(G_{Y}(y)) \approx y $, from which the objective can be expressed as:
			\begin{equation}
			\begin{aligned} 
			{\mathcal{L}}_{cyc}(G_{X},G_{Y})={{\mathbb{E}}_{x}}[\left\| {G_{Y}(G_{X}(x))-x} \right\|_1] + {{\mathbb{E}}_{y}}[\left\| {G_{X}(G_{Y}(y))-y} \right\|_1.
			\end{aligned} 
			\end{equation}
			Note that we still adopt a pix-level loss to satisfy backward cycle consistency, different from Kazemi \textit{et al}. \cite{Kazemi2018Unsupervised} who utilize a high-level perceptual loss for the cycle-consistency definition. 
			
			\textbf{Identity Perception Loss.}
			As illustrated in \cite{Zhang2018StackGAN}, only using the adversarial loss may lead to artifacts and training instabilities while synthesizing high-resolution images. Specifically, to add a strong supervision between the synthesized image and ground truth, we utilize a perception loss constructed by the face features relating to the activations of the fc7-layer of the VGG-16 \cite{Simonyan2014Very}. The identity perception loss $ {\mathcal{L}}_{ip} $ is defined as
			\begin{equation}\label{eq5}
			\begin{aligned} 
			{\mathcal{L}}_{ip}(G_{X},G_{Y})={{\mathbb{E}}_{x,y}}[\left\| {\Phi_p(G_{Y}(y))-\Phi_p(x)} \right\|_2^2] \\ + {{\mathbb{E}}_{x,y}}[\left\| {\Phi_s(G_{X}(x))-\Phi_s(y)} \right\|_2^2],
			\end{aligned} 
			\end{equation}
			where $\Phi_p$ and $\Phi_s$ are the recognition networks for photo and sketch.
			
			Unlike Wang \textit{et al.} \cite{Wang2018High}, who minimize the pixel-level difference between synthesized images and corresponding real samples, we employ a perceptual supervision for facial features for two reasons. 
			On the one hand, real sketches drawn by artists have distortions in the facial texture information and exaggerated facial characteristics in a certain style. Therefore, synthesizing images completely based on such hand-drawn sketches does not make sense. 
			Taking photo-to-sketch synthesis as an example, generated sketches have different texture distributions from hand-drawn sketches; using a pix-to-pix supervision of these two types of sketches is thus unreasonable.
			On the other hand, the training of CycleGAN requires common data augmentation techniques during data preprocessing, such as random cropping, flipping and resizing, which may cause pix-level supervision to be difficult to implement.
			
			\textbf{Identity Mapping Loss.} 
			To constraint the generator to be near an identity mapping when real samples from the target domain are provided as the input, we introduce a pixel-wise consistency between the input images and the generated images. We express this loss as:
			\begin{equation}
			\begin{aligned} 
			\mathcal{L}_{im}(G_{X},G_{Y})= \mathbb{E}_{x}\left[\left\|G_{X}(x)-x\right\|_{1}\right]
			+ \mathbb{E}_{y}\left[\left\|G_{Y}(y)-y\right\|_{1}\right] 
			\end{aligned}.
			\end{equation}
			
			\textbf{Full Objective.}
			Thus, the total objective function can be formulated as
			\begin{equation}\label{eq7}
			\begin{aligned}
			{\mathcal{L}}(G_X,G_Y,D_X,D_Y)={\mathcal{L}}_{GAN_X}+{\mathcal{L}}_{GAN_Y}+ \\ \lambda_{cyc}{\mathcal{L}}_{cyc} + \lambda_{ip}{\mathcal{L}}_{ip} + \lambda_{im}{\mathcal{L}}_{im},
			\end{aligned}
			\end{equation}
			where $ \lambda_{cyc} $, $ \lambda_{ip} $ and $ \lambda_{im} $ are parameters controlling the relative importance of every loss.
			The final objective can be expressed as a minmax problem:
			\begin{equation}\label{eq8}
			G _X ^ { * } , G _Y ^ { * } = \arg \min _ { G_{X} , G_{Y} } \max _ { D _ { X } , D _ { Y } } \mathcal { L } \left( G_{X} , G_{Y} , D _ { X } , D _ { Y } \right).
			\end{equation}

		\section{Mutual Optimization between the Synthesis and the Recognition Network} \label{Recognition Network}
			
\begin{figure}[htb]
				\centering
				\includegraphics[scale=0.49]{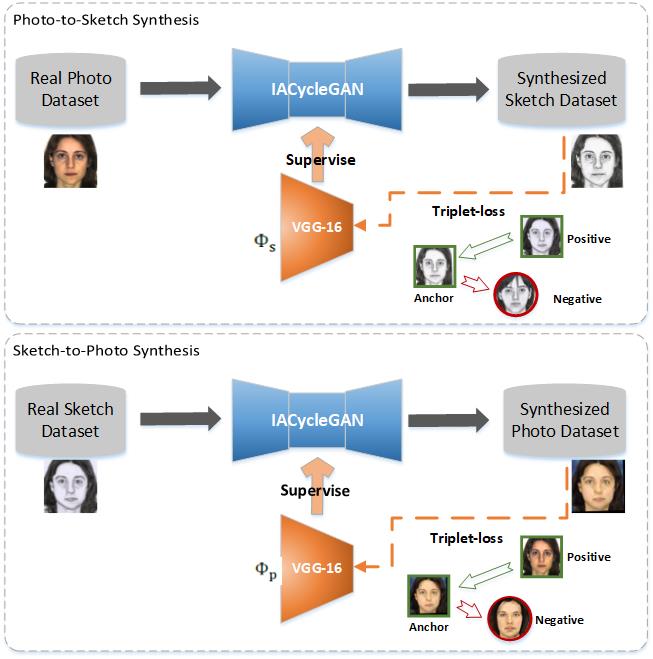}
				\caption{The mutual optimization process between the synthesis network, IACycleGAN and the recognition networks, $\Phi_p$ and $\Phi_s$. There are two image synthesis directions: photo-to-sketch synthesis (P2S) and sketch-to-photo synthesis (S2P). The synthesized dataset from IACycleGAN is used to retrain $\Phi_p$ or $\Phi_s$ with the triplet-loss, which aims to make each anchor image closer to the positive sample (corresponding fake or real image with the same identity) and far away from the negative samples (images from different people).}
				\label{fig4}
			\end{figure}

			Many similar studies adopting a perceptual loss use a fixed network to extract identity-specific information without a cyclic optimization \cite{Johnson2016Perceptual,Kazemi2018Unsupervised}. 
			Considering that the final goal is identity recognition and that an image-to-image translation model already exists, we can use the generated photo-sketch dataset from the basic synthesis network to retrain recognition networks. Thus the newly fine-tuned networks can be used as an additional supervision unit in training synthesis network.
			Note that the generated dataset mentioned above is constructed with all fake training images obtained from the synthesis model and the corresponding real training samples from the original dataset.
			After training IACycleGAN with new recognition networks, we obtain new fake photo and sketch datasets that can be applied to the fine-tuning process once again, which produce a new recognition model with better identity preserving ability.
			As shown in Figure \ref{fig4}, there are two models in our optimization system: the synthesis model ($ S $), used for producing fake images, and the recognition model ($\Phi_p$ and $\Phi_s$), which can be fine-tuned using the synthesized image dataset. Therefore, a mutual cyclic optimization process can be applied, as illustrated in Algorithm \ref{alg1}. Through the constant optimization between these two models, the recognition network is able to easily distinguish different subjects, and the synthesis network can be trained to generate images with more identity information.
			
			\begin{algorithm}[h]
				\caption{Mutual cyclic optimization procedure between the synthesis and the recognition network (IACycleGAN + VGGFace)} 
				\label{alg1}
				\footnotesize
				\begin{algorithmic}[1] 
					\REQUIRE~~\\
					Base synthesis network CycleGAN $S_0$;\\
					Base recognition network VGGFace $\Phi_0$;\\
					Real photo-sketch dataset $ \mathcal{D}_{real} $ ;
					\ENSURE~~\\ Optimized synthesis network $ S $ and recognition network $ \Phi $.
					\STATE Initializing $ i=0 $.
					\STATE Train $S_i$ on $ \mathcal{D}_{real} $ according to CycleGAN's objective function and generate fake photo-sketch dataset $ \mathcal{D}_{fake}^{(i)} $.
					\REPEAT 
					\STATE Fine-tune $\Phi_i$ using $ \mathcal{D}_{real} $ and $ \mathcal{D}_{fake}^{(i)} $ according to Eq. (\ref{eq9}) and obtain new recognition models $\Phi_{i+1}$ for photo and sketch.
					\STATE Train IACycleGAN $S_{i+1}$ on $ \mathcal{D}_{real} $ with $\Phi_{i+1}$ according to Eq. (\ref{eq8}) and generate fake photo-sketch dataset $ \mathcal{D}_{fake}^{(i+1)} $.
					\STATE $ i=i+1 $.
					\UNTIL{The image quality and recognition accuracy become stable.} 
				\end{algorithmic} 
			\end{algorithm}
			
			In the fine-tuning process, to make a real image closer to the corresponding fake sample from the same person and far away from the images of different people, the real image and the synthesized fake image should belong to the same class, and the labels from different people should be different.
			Therefore, we use a triplet loss \cite{Schroff2015FaceNet} to regularize the inter-class and intra-class distances, which aims to make the anchor image close to the positive sample and far from the negative samples.
			For each anchor image $ x_{i}^{a} $, where $ x_{i}^{p} $ is the corresponding fake or real positive image sample with the same identity and $ x_{i}^{n} $ is the negative sample from the images of a different person, we want to enforce the distance between $ x_{i}^{a} $ and $ x_{i}^{n} $ to be greater than it between $ x_{i}^{a} $ and $ x_{i}^{p} $. Thus, we add a triplet loss layer behind the fc8 layer in VGG-16 \cite{Simonyan2014Very}, and the loss is defined as
			\begin{equation}\label{eq9}
			\begin{aligned}
			{\mathcal{L}}_{tri} = \sum_{i}^{N}\left[\left\|\Phi\left(x_{i}^{a}\right)-\Phi\left(x_{i}^{p}\right)\right\|_{2}^{2}-\left\|\Phi\left(x_{i}^{a}\right)-\Phi\left(x_{i}^{n}\right)\right\|_{2}^{2}+\alpha\right]_{+},
			\end{aligned}
			\end{equation}
			where $\alpha$ is a margin enforced between positive and negative pairs and $\Phi$ is the recognition network.
			Moreover, to train a more robust feature extractor, we adopt the hard negative mining method in selecting the triplet samples \cite{Wang2015Unsupervised}, where only the hard negatives are chosen to calculate the loss, rather than all possible negative samples. Specifically, we perform gradient descent learning for the top $ K $ negatives with the highest losses calculated by the distance of the anchor image from all negative samples.

	\section{Experiments} \label{Experiments}
	
		In this section, we implement comparative analysis to verify the effectiveness of our proposed method for FPSS and FPSR tasks. Output images from the first synthesis task are used as the experimental datasets of our recognition work. According to Algorithm \ref{alg1}, our experiment was conducted by four steps: CycleGAN Synthesis $\rightarrow$ 1st VGG Fine-tuning $\rightarrow$ IACycleGAN Synthesis $\rightarrow$ 2nd VGG Fine-tuning. The synthesis experiments and final synthesis results in Step 1 and Step 3 are introduced in Section \ref{Photo-sketch Synthesis}; the fine-tuning experiments and recognition results in Step 2 and Step 4 are illustrated in Section \ref{recognition results}.
		
		\subsection{Datasets} \label{Datasets}
			There are only limited datasets for pairs of human-drawn sketches and face photos since it requires a great effort to collect face sketches. In this paper, the source of photo-sketch pairs is collected from two popular datasets: the Chinese University of Hong Kong (CUHK) Face Sketch Database (CUFS) \cite{Wang2009Face} and the CUHK Face Sketch FERET Database (CUFSF) \cite{Zhang2011Coupled}.
			
			The CUFS dataset \cite{Wang2009Face} consists of 606 faces in total: 188 faces from the CUHK student database \cite{Tang2002Face}, in which 88 pairs are taken as the training set and the remaining 100 pairs are used as the testing set; 123 faces from the AR database \cite{MARTINEZ1998The}, in which 80 pairs are used for the training and the remaining 43 pairs are used for testing; and 295 faces from the XM2VTS database \cite{Messer1999XM2VTSDB}, in which 100 pairs are chosen as the training dataset and the remaining 195 pairs are used as the testing set. Each face has one photo-sketch pair, and the sketch is drawn based on a photograph by an artist under normal lighting conditions as well as with neutral expressions.
			The CUFSF dataset \cite{Zhang2011Coupled} includes 1,194 faces from the FERET database \cite{Phillips1998The}, in which 250 faces are taken as the training set and the remaining 944 faces are taken as the testing set. Each person has a face photo with illumination changes and a face sketch drawn by an artist. This database is quite challenging since the photos are taken under different illumination conditions, and the sketches have numerous exaggerations compared to the photos.
			We employ the MTCNN method \cite{Zhang2016Joint} to perform face detection and alignment. Before the training process, all images are cropped to the standard size of 256$\times$256. 
		
		\subsection{Photo-sketch Synthesis} \label{Photo-sketch Synthesis}
			\textbf{Implementation Details:} 
			For the training of CycleGAN and IACycleGAN, all these synthesis networks are trained from scratch, instance normalization is used to achieve better stability and lower noise, and Adam with $ \beta_1 = 0.5$ and $ \beta_2 = 0.999$ is used. We flip the images horizontally with a probability of 0.5 for data augmentation. The learning rate is set to 0.0002 for the first 100 epochs and linearly decayed down to 0 for the next 100 epochs. Training takes about 10 hours with a single NVIDIA Titan Xp GPU. We empirically set the hyperparameters of the loss functions in Eq. (\ref{eq7}) as follows: $ \lambda_{cyc} = 10 $, $ \lambda_{ip} = 30000000 $, $ \lambda_{im} = 5 $.
			As illustrated in \cite{Zhu2017Unpaired}, to reduce network oscillation, an image buffer storing several generated images is adopted to update the discriminator rather than using the last image generated.
			
			Our final results of FPSS are obtained from the IACycleGAN synthesis in the optimization process. We compare three different networks (traditional CNN, conditional GAN and CycleGAN) with our proposed model from both qualitative and quantitative perspectives, therein focusing on the quality of the synthesized images. For the qualitative research, we employ visual observations. For the quantitative aspect, two measurements, the structural similarity index metric (SSIM) \cite{Wang2004Image} and the feature similarity index metric (FSIM) \cite{Zhang2011FSIM}, are used to evaluate structural and feature similarity between images.
			In addition, we test our trained synthesis network on other unseen datasets to prove the effectiveness of our framework.
			
			\subsubsection{Visual Observations} \label{Synthesis Results}
				
\begin{figure}[htb]
					\centering
					\subfigure[CUFS Dataset] {
						\centering
						\label{fig6a}     
						\includegraphics[scale=0.33]{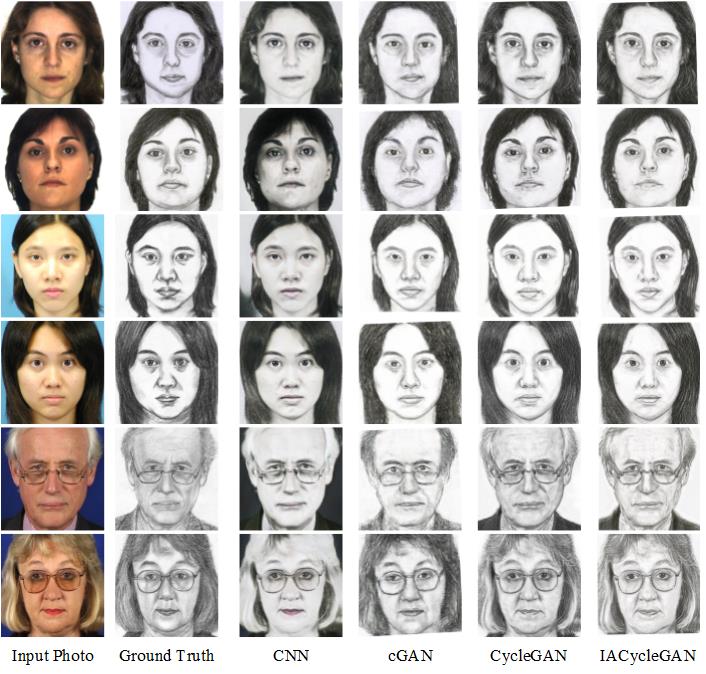}  
					} 
					\subfigure[CUFSF Dataset] { 
						\centering
						\label{fig6b}     
						\includegraphics[scale=0.33]{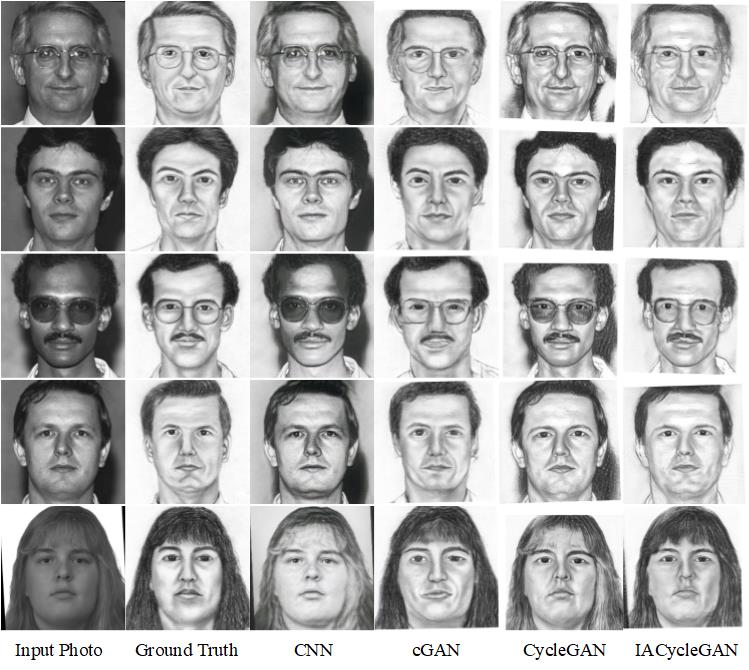}     
					} 
					\caption{Sample results for photo-to-sketch synthesis on the (a) CUFS and (b) CUFSF datasets generated by CNN \cite{Johnson2016Perceptual}, cGAN \cite{Isola2017Image}, CycleGAN \cite{Zhu2017Unpaired} and the proposed IACycleGAN.}     
					\label{fig6}     
				\end{figure}

\begin{figure}[htb]
					\centering
					\includegraphics[scale=0.35]{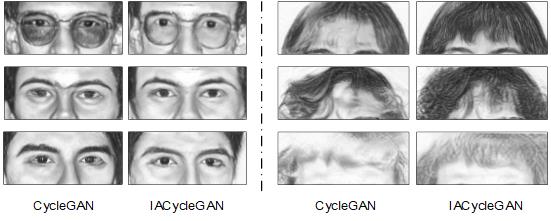}
					\caption{Comparisons of facial details of synthesized sketches for the baseline and IACycleGAN. The results obtained by IACycleGAN recovered more sketch-style details for the eyes and hair compared to the baseline method.}
					\label{fig66}
				\end{figure}

				Typical comparative results of P2S on the CUFS and CUFSF datasets are shown in Figure \ref{fig6}. 
				As observed from (a) and (b), the CUFSF dataset is significantly more challenging than CUFS because of the illumination changes of the photos and the over-exaggerated features of the sketches. 
				The traditional CNN method in the fast style transfer network \cite{Johnson2016Perceptual} leads to blurred results, where the synthesized images are not realistic and result in minimal information being learned from the sketch style. The other three methods can solve such problems due to the principles of adversarial loss. The results of cGAN (Pix2Pix) have some unacceptable distortions around the mouth and nose, whereas CycleGAN can produce clearer images. With the identity perception loss, some fuzzy edges near the eyes and nose can be eliminated, and the overall images seem to be clearer.
				Specifically, the facial details of the synthesized sketches in IACycleGAN tend to be nearer to the sketch style. For example, as shown in Figure \ref{fig66}, the eyes produced by IACycleGAN are more distinct and acceptable, and the glasses have been perfectly recovered. Moreover, the fringe and short hair around the forehead appears to be more similar to those of a hand-drawn sketch with fewer distortions in the representations of hair.
				
\begin{figure}[htbp]
					\centering
					\subfigure[CUFS Dataset] {
						\centering
						\label{fig7a}     
						\includegraphics[scale=0.32]{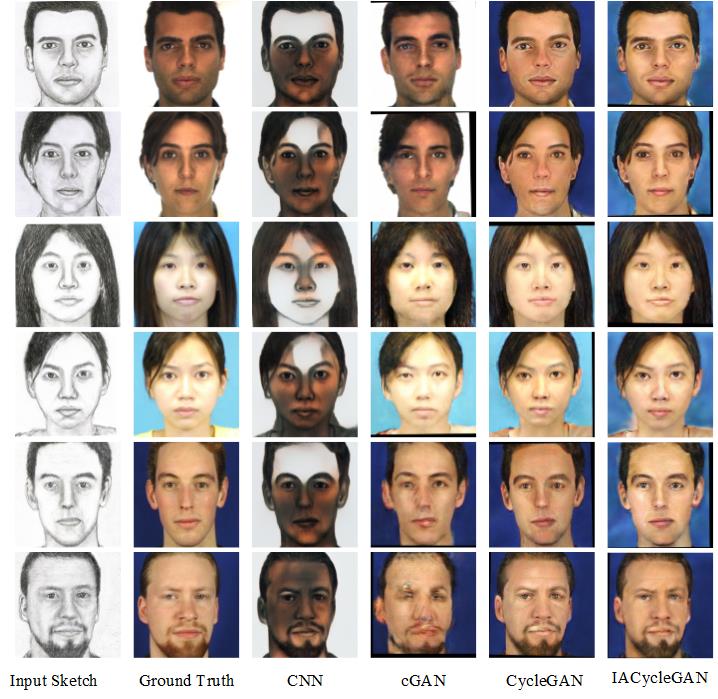}  
					} 
					\subfigure[CUFSF Dataset] { 
						\centering
						\label{fig7b}     
						\includegraphics[scale=0.35]{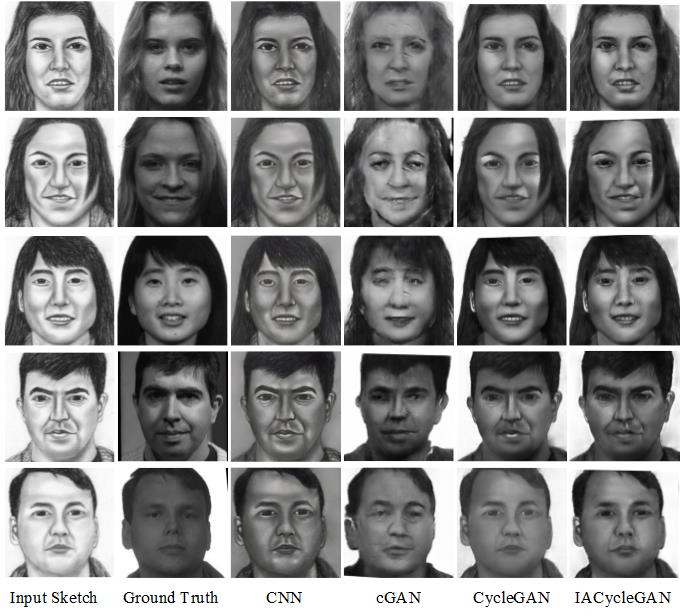}     
					} 
					\caption{Sample results for sketch-to-photo synthesis on the (a) CUFS and (b) CUFSF datasets generated by CNN \cite{Johnson2016Perceptual}, cGAN \cite{Isola2017Image}, CycleGAN \cite{Zhu2017Unpaired} and the proposed IACycleGAN.}     
					\label{fig7}     
				\end{figure}

\begin{figure}[htbp]
					\centering
					\includegraphics[scale=0.34]{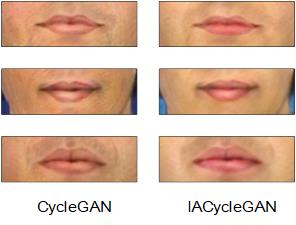}
					\caption{Comparisons of facial details of synthesized photos for baseline and IACycleGAN. The lips of the synthesized photos from IACycleGAN contain realistic characteristics, therein being more lifelike and similar to a real face.}
					\label{fig77}
				\end{figure}

				Figure \ref{fig7} shows the results of S2P on the two datasets. The synthesized photos produced by the CNN method are still not realistic, and suffer color distortions a lot on the CUFS dataset. CGAN produces highly blurred images compared to other GAN methods. 
				The CycleGAN model seems effective but introduces undesirable artifacts around the nose. Compared to the results of CycleGAN, IACycleGAN generates images with more real facial details, such as details about the eyes and lips, which make the fake image look more like a face photo rather than a sketch. Although these artifacts cannot be completely eliminated with the identity perception loss, the overall quality of the images is improved, and some lightness distortions of CUFSF are partially corrected.
				Similarly, the facial details of the synthesized photos are shown in Figure \ref{fig77}. Our IACycleGAN model can effectively generate red and real lips, while CycleGAN produces some local texture of the pencil drawing around the edges of the lips. 
				
			\subsubsection{Quantitative Measurements}
			
				The SSIM (structural similarity index) \cite{Wang2004Image} is an index for measuring the perceptual metric difference between a synthesized image and its corresponding ground-truth image. It defines the structure information as an attribute in a scene that is independent of brightness and contrast with respect to image composition.
				The FSIM (feature similarity index) \cite{Zhang2011FSIM} is a type of image quality assessment that evaluates the image denoising performance and has a similar function to SSIM.
				Thus, these two measurements can be used to produce a better estimation of the quality of synthesized images as well as the performance of different synthesis methods.
				
				\begin{table}[htbp]
					\footnotesize 
					\centering
					\begin{tabular}{ccccc}
						\hline
						\multicolumn{2}{c}{} & cGAN\cite{Isola2017Image} & CycleGAN\cite{Zhu2017Unpaired} & IACycleGAN\\
						\hline\hline
						\multirow{2}{*}{sketch} & CUFS($\%$) & 63.72 & 64.52 & \textbf{64.95} \\
						& CUFSF($\%$) & 49.39 & 57.05 & \textbf{57.64} \\
						\hline
						\multirow{2}{*}{photo} & CUFS($\%$) & 69.26 & 70.70 & \textbf{71.22} \\
						& CUFSF($\%$) & 57.09 & 61.50 & \textbf{62.50} \\
						\hline
					\end{tabular}
					\caption{Comparison of average SSIM scores of two synthesis directions based on the CUFS and CUFSF datasets for three GAN-based methods.}
					\label{tab1}
				\end{table}
				
				\begin{table}[htbp]
					\footnotesize 
					\centering
					\begin{tabular}{ccccc}
						\hline
						\multicolumn{2}{c}{} & cGAN\cite{Isola2017Image} & CycleGAN\cite{Zhu2017Unpaired} & IACycleGAN\\
						\hline\hline
						\multirow{2}{*}{sketch} & CUFS($\%$) & 73.22 & 73.74 & \textbf{74.01} \\
						& CUFSF($\%$) & 69.41 & 68.67 & \textbf{70.05} \\
						\hline
						\multirow{2}{*}{photo} & CUFS($\%$) & 75.77 & 76.24 & \textbf{76.52} \\
						& CUFSF($\%$) & 70.21 & 72.41 & \textbf{75.47} \\
						\hline
					\end{tabular}
					\caption{Comparison of average FSIM scores of two synthesis directions based on the CUFS and CUFSF datasets for three GAN-based methods.}
					\label{tab2}
				\end{table}
				
				The result images from the three GAN-based methods are tested using these two measurements. Table \ref{tab1} shows the comparative results of the SSIM, and Table \ref{tab2} presents the results of the FSIM. From these two tables, we can see that IACycleGAN provides higher scores compared to the baseline GAN models in terms of both criteria. 
				Simultaneously, Table \ref{tab3} shows comparison results of the average SSIM scores (\%) of the LLE \cite{Liu2005A}, SFS \cite{Gao2012Face}, SSD \cite{Song2014Real-Time}, MRF \cite{Wang2009Face}, MWF \cite{Zhou2012Markov}, Fast-RSLCR \cite{Wang2017Random}, FCN \cite{Zhang2015End-to-End}, LR \cite{Wang2017Data-driven}, and IACycleGAN models on the two datasets. We can observe that our proposed method outperforms all other competing models. 
				In general, by fusing the perceptual-level supervision in the baseline model based on two recognition networks, the proposed method can learn a more discriminative feature representation in terms of both qualitative and quantitative aspects.
				
				\begin{table}[htbp]
					\centering
					\resizebox{\textwidth}{10.5mm}{
						\begin{tabular}{cccccccccc}
							\hline
							& LLE & SFS & SSD & MRF & MWF & Fast-RSLCR & FCN & LR\ & IACycleGAN\\
							\hline\hline
							CUFS & 52.58 & 51.90 & 54.20 & 51.32 & 53.93 & 55.42 & 52.14 & 54.20 & \textbf{64.95} \\
							CUFSF & 41.79 & 42.11 & 44.09 & 37.24 & 42.99 & 44.55 & 36.22 & 36.65 & \textbf{57.64} \\
							\hline
						\end{tabular}}
					\caption{Comparison of the average SSIM scores (\%) of the sketch synthesis for several popular methods based on the CUFS and CUFSF datasets.}
					\label{tab3}
				\end{table}
			
			\subsubsection{Results on Unseen Datasets}
\begin{figure}[htbp]
					\centering
					\includegraphics[scale=0.37]{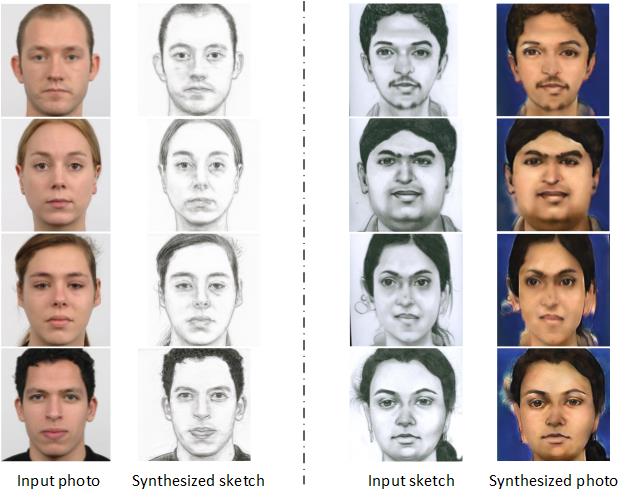}
					\caption{Sample results of photo-to-sketch synthesis (left) on RaFD and sketch-to-photo synthesis (right) on IIIT-D using trained IACycleGAN model.}     
					\label{fig11}     
				\end{figure}

				To validate the effectiveness of this method, we test our trained IACycleGAN on other unseen datasets without prior knowledge from testing data. We utilize the network trained on CUFS, since the images of this dataset look more common and have no special condition compared to others. For photo-to-sketch synthesis, we tested the photos from RaFD \cite{Langner2010Presentation}, and for sketch-to-photo synthesis, we tested the viewed sketches from the IIIT-D Sketch Database \cite{Bhatt2012Memetic}. Figure \ref{fig11} shows some example results, where the synthesized images look quite acceptable though no training on input datasets.
				
		\subsection{Photo-sketch Recognition} \label{recognition results}
			
			\textbf{Implementation Details:} 
			The fine-tuning process is implemented on Caffe. All networks are trained using $ momentum=0.9, weight_{decay}=0.0002 $.
			For first fine-tuning on the basic VGGFace, we adopt $ step $ learning policy with $ stepsize=100, gamma=0.96, base_{lr}=0.3 $, and 600 iterations are used for photo recognition network and 1600 iterations for sketch recognition network.
			For second fine-tuning on the last fine-tuned VGG, we adopt $ step $ training policy with $ stepsize=200, gamma=0.96, base_{lr}=0.01 $, and 2000 iterations are used for both photo and sketch recognition networks.
			We set $ K=4 $ and $ \alpha = 0.1$ for the parameters of triplet loss layer. Training takes about 30 hours with a single NVIDIA Titan Xp GPU.
			
			\begin{figure}[htbp]
				\centering
				\subfigure[Photo matching method.] {
					\centering
					\label{fig5a}     
					\includegraphics[scale=0.48]{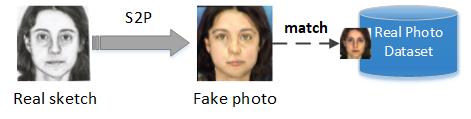}  
				} 
				
				\subfigure[Sketch matching method.] { 
					\centering
					\label{fig5b}
					\includegraphics[scale=0.48]{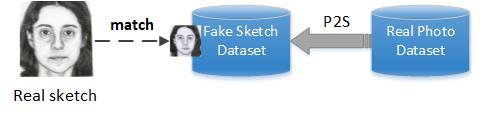}     
				} 
				\caption{Two approaches for photo retrieval application in matching a real query sketch to a photo dataset.}     
				\label{fig5}     
			\end{figure}
			
			As seen from Figure \ref{fig5}, there are two approaches to retrieving real face photos based on a query sketch: photo matching method and sketch matching method. 
			In the photo matching method, sketch-to-photo synthesis (S2P) is first implemented on the query sketch; then, the generated photo is used for matching on the real photo dataset, where the photo matching rate needs to be computed. 
			In the sketch matching method, all photos in the dataset will first be transformed into sketches, and then, the query sketch is matched to the generated fake samples; at last, the sketch matching rate is computed.
			In our experiment, we implement both two matching methods and compare the results with other synthesis-based recognition techniques.
			
			\subsubsection{Recognition for GAN-Based Methods}\label{Recognition Rate}
				
				Table \ref{tab4} shows the recognition results of three GAN-based synthesis models, while the recognition model is VGG-Face \cite{Parkhi2015Deep}. 
				We also implement a non-synthesis-based method, which means directly matching between two domains without any synthesis process. It can be observed that synthesis-based methods show a better recognition ability, and our IACycleGAN performs better than other baseline models for both photo and sketch recognition on the two datasets.
				Moreover, by using score fusion to combine the recognition rates of two matching methods, the performance can be further improved.
				
				\begin{table}[htbp]
					\footnotesize 
					\centering
					\begin{tabular}{cc|c|c|c|c}
						\hline
						\multicolumn{2}{c|}{Synthesis model} & none & cGAN & CycleGAN & IACycleGAN \\
						\hline
						\multicolumn{2}{c|}{Recognition model} & \multicolumn{4}{c}{VGG-Face} \\
						\hline\hline
						\multirow{3}{*}{CUFS} & sketch-matching & \multirow{2}{*}{94.97} & 97.63 & 98.82 & \textbf{100.00 } \\
						& photo-matching & & 81.95 & 96.45 & \textbf{98.82} \\
						\cline{2-6}
						& fusion & 94.97 & 99.41 & 98.82 & \textbf{100.00} \\
						\hline\hline
						\multirow{3}{*}{CUFSF} & sketch-matching & \multirow{2}{*}{57.94} & 41.21 & 68.75 & \textbf{70.87} \\
						& photo-matching & & 28.07 & 59.00 & \textbf{64.94} \\
						\cline{2-6}
						& fusion & 57.94 & 57.94 & 73.52 & \textbf{81.36} \\
						\hline
					\end{tabular}
					\caption{Comparison of the rank-1 recognition accuracy ($\%$) on the CUFS and CUFSF datasets for three GAN-based methods. Both sketch and photo recognition methods are implemented, and fusion means using score fusion to combine two matching methods.}
					\label{tab4}
				\end{table}
				
\begin{figure}[htb]
					\centering
					\includegraphics[scale=0.45]{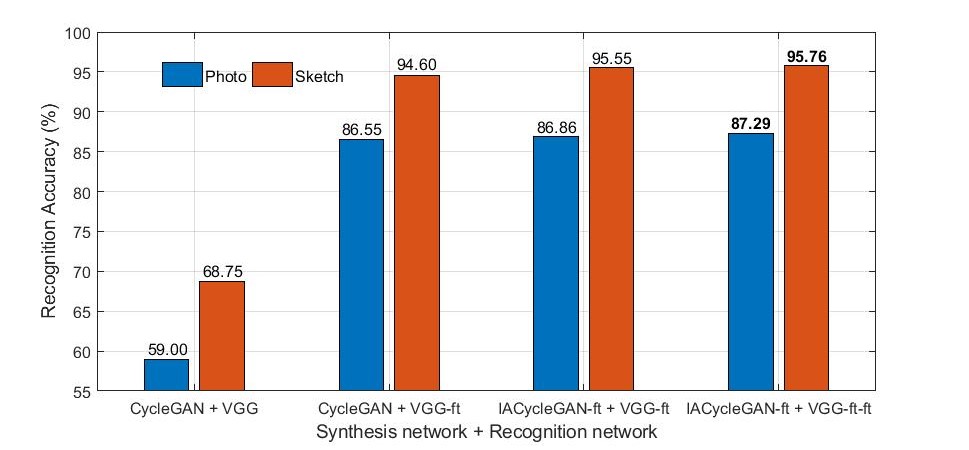}
					\caption{The results of iterative optimization to FPSR on the CUFSF dataset. VGG-ft is the fine-tuned VGG recognition network, IACycleGAN-ft means using the fine-tuned recognition network VGG-ft to supervise the synthesis network IACycleGAN, and VGG-ft-ft means fine-tuned from the last recognition network, VGG-ft. 
						Every synthesis network has one generated photo-sketch dataset corresponding to two accuracies computed by original recognition network $\Phi_{i}$ and the newly fine-tuned recognition network $\Phi_{i+1}$, as illustrated in Algorithm \ref{alg1}.
					}
					\label{fig10}
				\end{figure}

				In our optimization model, every new synthesized dataset can be tested directly or used to fine-tune the last recognition network and then compute matching rates on the new recognition network. Thus, two recognition results are obtained for one synthesis model with these two recognition model, as shown in Figure \ref{fig10}.
				We can also observe that with further optimization of this cyclic system, higher recognition accuracy is obtained, validating the efficiency and effectiveness of the proposed cyclic optimization process.
				Specifically, due to the simplicity of the CUFS dataset, the recognition accuracy approaches 100\% for the first fine-tuned result, as shown in Table \ref{tab4}. In other words, image data may be somewhat distorted on CUFS by multiple iterations and over-sampling; thus, further cyclic fine-tuning is only applied to the CUFSF dataset in our experiment. 
				
			\subsubsection{Comparisons to Other Methods}
				
				In addition, we compare our final recognition results with several state-of-the-art FPSR methods using conventional face recognition algorithms. 
				For the CUFS dataset, 150 synthesized images with corresponding ground truth are randomly selected for classifier training, and the testing set contains the remaining 188 image pairs. And for CUFSF, 300 fake-real image pairs are selected for training, and the remaining 644 pairs are used for testing.
				For sketch-matching method, we adopt nullspace linear discriminant analysis (NLDA) \cite{Chen2000A} as the basic recognition technique. 
				The comparison results for sketch-matching are shown in Table \ref{tab5}, therein referring to \cite{Akram2018A}, who uses the same training protocols as our experiment.
				Meanwhile, Zhang \textit{et al.} \cite{Zhang2019Dual} adopt the eigenface recognition technique to compare several synthesis models based on photo-matching. Since we have similar evaluation settings as their experiment, we implement eigenface recognition for photo matching method, the results of which are summarized in Table \ref{tab6}.
				
				\begin{table}[h]
					\footnotesize
					\centering
					\begin{tabular}{ccc}
						\hline
						Datasets & CUFS & CUFSF \\
						\hline\hline
						LLE \cite{Liu2005A} & 90.61 & 61.09 \\
						SFS \cite{Gao2012Face} & 89.60 & 72.48 \\
						SSD \cite{Song2014Real-Time} & 90.00 & 70.68 \\
						MRF \cite{Wang2009Face} & 87.34 & 45.26 \\
						MWF \cite{Zhou2012Markov} & 92.10 & 74.05 \\
						Fast-RSLCR \cite{Wang2017Random} & 98.43 & 72.77 \\
						FCN \cite{Zhang2015End-to-End} & 96.49 & 69.80 \\
						LR \cite{Wang2017Data-driven} & 91.67 & 25.93 \\
						GAN \cite{Isola2017Image} & 93.48 & 71.44 \\
						IACycleGAN (ours) & 98.40 & 74.53 \\
						\hline
						Mutual Optimization (Algorithm \ref{alg1}) & \textbf{100.00} & \textbf{95.76} \\
						\hline
					\end{tabular}
					\caption{Comparison of the rank-1 recognition accuracy (\%) for sketch matching method on the CUFS and CUFSF datasets. All other synthesis models are using the NLDA recognition technique, and we apply both NLDA and the proposed mutual optimization recognition method to our IACycleGAN synthesis model.}
					\label{tab5}
				\end{table}
				
				\begin{table}[htbp]
					\footnotesize
					\centering
					\begin{tabular}{ccc}
						\hline
						Datasets & CUFS & CUFSF \\
						\hline\hline
						LLE \cite{Liu2005A} & 85.0 & -- \\
						MRF \cite{Wang2009Face} & 85.7 & -- \\
						MWF \cite{Zhou2012Markov} & 86.3 & -- \\
						SNS-SVR \cite{Wang2011Face} & 85.0 & -- \\
						TFSPS \cite{Wang2013Transductive} & 84.3 & -- \\
						DR-GAN \cite{Tran2017Disentangled} & 83.7 & -- \\
						Dual-Transfer \cite{Zhang2019Dual} & 86.3 & -- \\
						IACycleGAN (ours) & 93.62 & 42.55 \\
						\hline
						Mutual Optimization (Algorithm \ref{alg1}) & \textbf{98.82} & \textbf{87.29} \\
						\hline
					\end{tabular}
					\caption{Comparison of the rank-1 recognition accuracy (\%) for photo matching method on the CUFS and CUFSF datasets. All other synthesis models are using the Eigenface recognition technique, and we apply both eigenface and the proposed mutual optimization recognition method to our IACycleGAN synthesis model.}
					\label{tab6}
				\end{table}
			
				In addressing the challenges facing photo matching method on the CUFSF dataset, minimal recognition data can be obtained in this direction. Accordingly, a summary of other photo-matching-based results is only provided on the CUFS dataset, as summarized in Table \ref{tab6}.
				From the comparison results in Table \ref{tab5} and Table \ref{tab6}, our synthesis model, IACycleGAN, outperforms all other competitors with the same recognition algorithm.
				We can observe that the recognition results of CUFS are much better than CUFSF and approach to 100\% when using the basic recognition method on IACycleGAN. Thus, there remains enough upside for the recognition results of CUFSF, which accordingly realized a significant increase compared to CUFS.
				Moreover, we can also observe that the proposed recognition method (IACycleGAN + VGG) achieves the best matching accuracy by using a mutual optimization technique between the IACycleGAN synthesis model and the recognition model.

	\section{Conclusion} \label{Conclusion}
	
		In this paper, we propose Identity-aware CycleGAN IACycleGAN for achieving photo-sketch synthesis with better identity preserving ability and apply a mutual optimization method to explore FPSS and FPSR problems simultaneously. 
		For FPSS, the proposed IACycleGAN adopts a new perceptual supervision which is constructed with the deep facial features from a pre-trained recognition network.
		This synthesis model shows better performance compared to some recent popular image-to-image translation models for both visual observations and two quantitative measurements (SSIM and FSIM).
		For FPSR, the recognition network can be re-trained using the feedback from FPSS, a newly synthesized image dataset from IACycleGAN; the synthesis process can then be trained with the fine-tuned recognition model once again. Thus, a mutual cyclic optimization process is formed between the synthesis model and the recognition model, improving both FPSS and FPSR.	
		This method is proven to greatly enhance the recognition accuracy compared to baseline models and other state-of-the-art methods for both sketch and photo matching.
		All results are evaluated on two popular datasets, CUFS \cite{Wang2009Face} and CUFSF \cite{Zhang2011Coupled}, and the proposed method achieved considerable improvements in terms of both image quality and photo-sketch matching accuracy.	
		Our work explores the relation between the synthesis model and the recognition model, from which we plan to improve the FPSS and FPSR on a more common dataset by strengthening and stabilize the connection.
	
	\section{Acknowledgements}
		This work was partially supported by the National Natural Science Foundation of China under Grant Nos. 61573068 and 61871052.

	\section{References}
	{\footnotesize
		\bibliographystyle{IEEEtran}
		\bibliography{sketch}
	}
	
\end{document}